%% file: naacl2021.tex
\pdfoutput=1
\documentclass[11pt]{article}
\usepackage{authblk}
\usepackage[]{naacl2021}

\usepackage{times}
\usepackage{latexsym}

\usepackage[T1]{fontenc}
\usepackage[utf8]{inputenc}

\usepackage{microtype}

\usepackage{amsmath}
\usepackage{graphicx}
\usepackage{subfigure}
\usepackage{booktabs}
\graphicspath{ {./figures/} }
\usepackage{makecell}

\title{Language Scaling for Universal Suggested Replies Model}

\author[1]{Qianlan Ying\thanks{\hspace{0.15cm}Both authors contributed equally in the paper.}\hspace{0.15cm}}
\author[2]{Payal Bajaj\protect\footnotemark[1]\hspace{0.15cm}}
\author[2]{Budhaditya Deb}
\author[1]{Yu Yang} 
\author[3]{Wei Wang\thanks{\hspace{0.15cm}Work performed at Microsoft Research.}\hspace{0.15cm}}
\author[1]{\\ Bojia Lin}
\author[2]{Milad Shokouhi}
\author[2]{Xia Song}
\author[1]{Yang Yang}
\author[1]{Daxin Jiang}
\affil[1]{Microsoft, Beijing, China}
\affil[2]{Microsoft, Bellevue, Washington, USA}
\affil[3]{Qualtrics, Seattle, Washington, USA}
\affil[ ]{\texttt {\{qiying,Payal.Bajaj,Budha.Deb,yanyu\}@microsoft.com}}
\affil[ ]{\texttt {\{bojial,milads,xiaso,yayan,djiang\}@microsoft.com}}
\affil[ ]{\texttt {tskatom@gmail.com}}

\usepackage{multirow}
\begin{document}

\maketitle
\input{tex/abstract}

\section{Introduction}
\input{tex/introduction}

\section{Core SR Model}
\input{tex/sr_model}

% \begin{figure}[htbp!]
%     \centering
%         \subfigure[]
%             {
%                 \begin{minipage}[htbp]{1.0\linewidth}
%                 \centering
%                 \includegraphics[width=0.9\textwidth]{figures/architecture.png}
%                 %\includegraphics[width=1.0\textwidth]{figures/diagram_02.emf}

%                 \label{fig:b}
%                 \end{minipage}
%             }
%         \subfigure[]
%             {
%                 \begin{minipage}[htbp]{1.0\linewidth}
%                 \centering
%                 \includegraphics[width=1.0\textwidth]{figures/training_flow.png}
%                 \label{fig:a}
%                 \end{minipage}
%             }%
% \caption{(a) Matching model architecture with symmetric loss and TLM/MLM cross-entropy loss. 
% % loop of universal model. The in-domain training data is restricted to 3 clusters with the model trained and exported sequentially as EUR->NAM->LRL. 
% (b) Multi-task continual training loop for EUR->NAM->LRL clusters.
% % Messages and responses are encoded by two independent encoders with exactly same architecture. 
% % They are initialized with XLM-R like cross-lingual pretrained language encoder then fine-tuned separately. At inference, only Message Encoder is used to encode input messages, while fixed responses are encoded offline. Finally, each M-R matching score is computed by cosine similarity.
% }
% \label{fig1}
% \end{figure}

\section{Universal SR Model\label{universalmodel}}

\input{tex/universal_model}

\section{Experiments and Results}
\input{tex/experiments}

\section{Conclusions}
This paper presents our approach of scaling automated suggested replies with one universal model. Faced with compute resource and data privacy constraints, we propose a multi-task continual learning framework with auxiliary tasks, and data augmentation with adapter-based model architecture. The universal model in production saves significant compute resources and model management overhead, while allowing us to train across regional data boundaries. In addition, the process allows us to \textit{cold-start} in new markets even when no supervised data exists. Based on the promising offline and online results, we have deployed the model in several languages and plan to extend  the process for 20 languages around the world.

% \section*{Acknowledgements}

\bibliography{naacl2021}
\bibliographystyle{acl_natbib}
\appendix

% \section{Example Appendix}
% \label{sec:appendix}

% This is an appendix.

\end{document}

%% file: tex/abstract.tex
\begin{abstract}
We consider the problem of scaling automated suggested replies for Outlook email system to multiple languages. Faced with increased compute requirements and low resources for language expansion, we build a single \textit{universal} model for improving the quality and reducing run-time costs of our production system. However, restricted data movement across regional centers prevents joint training across languages. To this end, we propose a multi-task \textit{continual learning} framework, with auxiliary tasks and language adapters to learn universal language representation across regions. The experimental results show positive cross-lingual transfer across languages while reducing \textit{catastrophic forgetting} across regions. Our online results on real user traffic show significant gains in CTR and characters saved, as well as 65\% training cost reduction compared with per-language models. As a consequence, we have scaled the feature in multiple languages including low-resource markets.
\end{abstract}

%% file: tex/introduction.tex
Automated suggested replies or smart replies (SR) assist users to quickly respond with a short, generic, and relevant response, without users having to type in the reply. SR is an increasingly
popular feature in many commercial applications such as Gmail, Outlook, Skype, Facebook Messenger, Microsoft Teams, and Uber \cite{Kannan2016,HendersonASSLGK17,shang2015neural, Deb2019DiversifyingRS,uberSR}. While the initial versions of this feature mostly targeted English users, making it available in multiple languages and markets is important not only from the perspective of product expansion but also from a linguistic inclusivity point of view.

In this paper we consider the problem of rapid scaling of the SR feature to multiple languages for Outlook. To develop such a system at production scale, we are faced with the following challenges.
% In these applications, one or several short, generic and relevant replies are recommended given one short message to assist user with quick communication by one click without typing.
%\includegraphics{sr_example}

% motivation for universal Smart Reply: challenges in language scaling

% The success of such NLP applications in En-US market has led to language scaling to international markets.
% Challenges of building per-language separate models
% 1. A number of per-models
% To achieve the ambitious goal of extending the feature to global markets. 
- \textbf{Model management}: Language scaling increases the effort of training, deploying, and managing per-language models, which needs to be replicated for each language. In addition, one model per language increases the storage and compute requirements for the production servers, which can increase costs and occurrences of run-time issues.

% However, agility is a critical key. However, the biggest challenge of scaling globally is the linearly increasing effort of building per-language models. Similar processes of model training and serving have to be replicated multiple times %TO-DO: Refine the phrase
% , which is computationally costly and time-expensive. A number of models are required to be maintained for corresponding markets, which is also memory consuming. 

% 2. Lack of data
- \textbf{Data constraints}: Developing models at production quality requires considerable effort in data collection and management. Due to regional market share and infrastructure constraints, rich and domain-specific data may not be available for all languages. 
% independent models per language requires Each, new locales are also faced with challenge of data collection. Typically, only a few of markets are with rich data, while others are not, due to market share and government policies. Current strategies of building separate models can't leverage existing data in high-resource markets to achieve high accuracy in markets lack of data or even without any data.  % TO-DO: Refine the phrase
% meaningful to investigate how to leverage data from a few locales and applies across all locales

% 3. Data region constraints
- \textbf{Data privacy and security policies}: 
% All user data is processed \textit{eyes-off}, which means that no developer has any direct access to the data. 
Regional policies enforce data to be located in corresponding regions. For example, Spanish and Portuguese data are stored in North American (NAM) clusters while French data is stored in European (EUR) clusters. Data movement across regions is not allowed and this prevents leveraging commonly used multi-lingual co-training methods which require all the data stored to be in the same place.

% from is mandatory and raises a new challenge for model building in many business scenarios. Except that the data is collected without eyes access, even for the mature markets that are capable of collecting enough native data, it is required to keep user data scattered in corresponding regions. 
 % TO-DO: Investigate compliance policy for refinement % common method of multi-lingual model. 

% Our proposed method and contribution
% framework: universal model + technical methods for universal model ( pre-trained model + continual learning + multi-task training) + technical methods for low-resource markets (MT data augmentation)
% The language scaling approach in this paper is motivated from these challenges. 
To reduce the cost of model management, we propose to build a single \textit{universal} SR model, capable of serving multiple languages and markets. To overcome data constraints, we propose to use \textit{augmentation} with machine-translated (MT) data for languages without supervised data. 
% doesn't require supervised data in all languages augmented with machine-translated (MT) data. 
To overcome privacy constraints, we propose a \textit{continual} learning framework, where the model is trained sequentially across regions. To alleviate \textit{catastrophic forgetting} \citep{french1999catastrophic,mccloskey1989catastrophic} in the continual learning process, 
we reinforce the universal properties via multi-task learning approach with public task-agnostic data, and an adapter-based model architecture that leverages domain-specific SR data and MT data.

%Our model is based on the standard matching model architecture commonly used in SR systems \cite{HendersonASSLGK17,Deb2019DiversifyingRS}. We extend this to a universal setting where we fine-tune pre-trained universal text encoders. 
% (Repetitive with last paragraph) For low-resource languages, we augment the data using machine translation (MT) service from English message-reply pairs. Additionally we use publicly available multi-lingual data for auxiliary tasks in our continual learning framework.

Our experimental results followed with improvements shown on real user traffic illustrate the effectiveness of the approach. As a consequence, we have rapidly scaled  the  feature  in  several  languages  including low-resource markets. Multi-lingual training for universal models is often very tricky to work in practice  (especially with our data constraints). Thus, we demonstrate a significant accomplishment of a multi-lingual SR system running at production scale on millions of users, which saves resources while improving performance.

%% file: tex/sr_model.tex
The SR feature is similar to open-domain chat-bots and task-oriented conversational agents, \cite{zhou2020design, Henderson2019b,fadhil2019designing,xu2017new,okuda2018ai,kopp2018conversational}. In terms of usage, SR is closer to the latter, in that it assists users to complete a reply, instead of continuing an open-ended dialog. Following commonly used IR-based models in commercial SR applications \cite{henderson2017efficient, Deb2019DiversifyingRS}, we use a dual encoder matching model for our SR system.

The matching model has two parallel encoders projecting input message and corresponding reply into a common representation space. Different encoders such as feed-forward and BiLSTM layers can be used here \cite{HendersonASSLGK17, Deb2019DiversifyingRS}. More recently, \cite{devlin2018bert,liu2019roberta,yang2019xlnet,henderson2019convert,Henderson2019b} show considerable improvements with transformer-based pre-trained models. Our English SR model uses a BERT equivalent \cite{devlin2018bert} encoder, while our mono-lingual baselines in other languages use BiLSTM encoders. 

% For mono-lingual models in other languages we use BiLSTM encoder as baselines. 

% Due to the challenges to align multi-user conversations without direct access, we filter the data to just one-to-one conversations. The m-r pairs are used for training a dual encoder architecture described next.

% During data pre-processing, only email body is extracted with personal sensitive information replaced by masks with special tokens such as \textit{\#Name\#} and \textit{\#PhoneNumber\#}.

% % \subsection{Matching Model}
% The matching model is composed of two parallel text encoders which project input message and corresponding reply into a common representation space $[\phi(m_{i}), \phi(r_{i})]$. 
% There are multiple candidates to construct such a text encoder. For example, \cite{HendersonASSLGK17} uses a stack of feed-forward layers  while \cite{Deb2019DiversifyingRS}, uses a shared embedding layer with stacked bi-directional LSTM layers. 
% Exploration on ranking tasks with pre-trained language models is also an emerging trend \cite{qiao2019understanding}. 

% Similarly, pre-trained models are widely adopted in multi-lingual systems to extract the language-agnostic features. We use such a pre-trained multi-lingual encoder which will be described in Section \ref{universalmodel}.

The model is trained on one-on-one message-reply (m-r) pairs from commercial email data. We minimize the symmetric loss function. It is a modified softmax on dot products between m-r encoding in equation \ref{symloss} where  $s_{i,j}=e^{\phi(m_{i}) \cdot \phi(r_{j})}$. As described in \cite{Deb2019DiversifyingRS}, it was shown to improve the relevance by targeting at bi-directional \textit{conversational} constraints.

{\small
	\begin{equation}\label{symloss}
	    p(m_{i}, r_{i})\\ =\frac{s_{i,i}} 
         {\sum_{j}{s_{i,j}} 
         + \sum_{k}{s_{k,i}} 
         - s_{i,i}}
	\end{equation}
}

% \subsection{Response Set Generation}
\begin{figure}[htb]
    \centering
        \subfigure[]
            {
                \begin{minipage}[htbp]{1.0\linewidth}
                \centering
                \setlength{\abovecaptionskip}{0.cm}
                \includegraphics[width=0.9\textwidth]{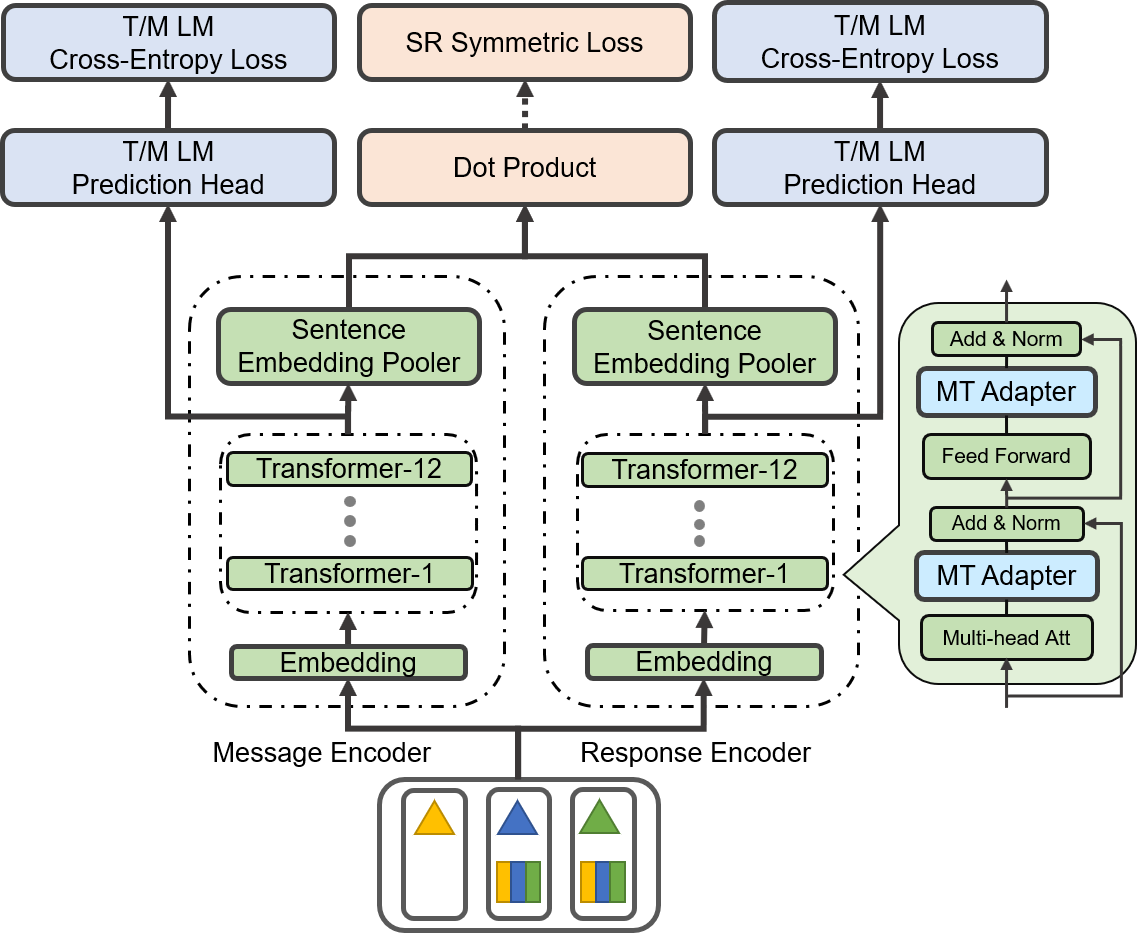}
                \label{fig:a}
                \end{minipage}
            }\hspace{-30mm}
        \subfigure[]
            {
                \begin{minipage}[htbp]{1.0\linewidth}
                \centering
                 \includegraphics[width=0.9\textwidth]{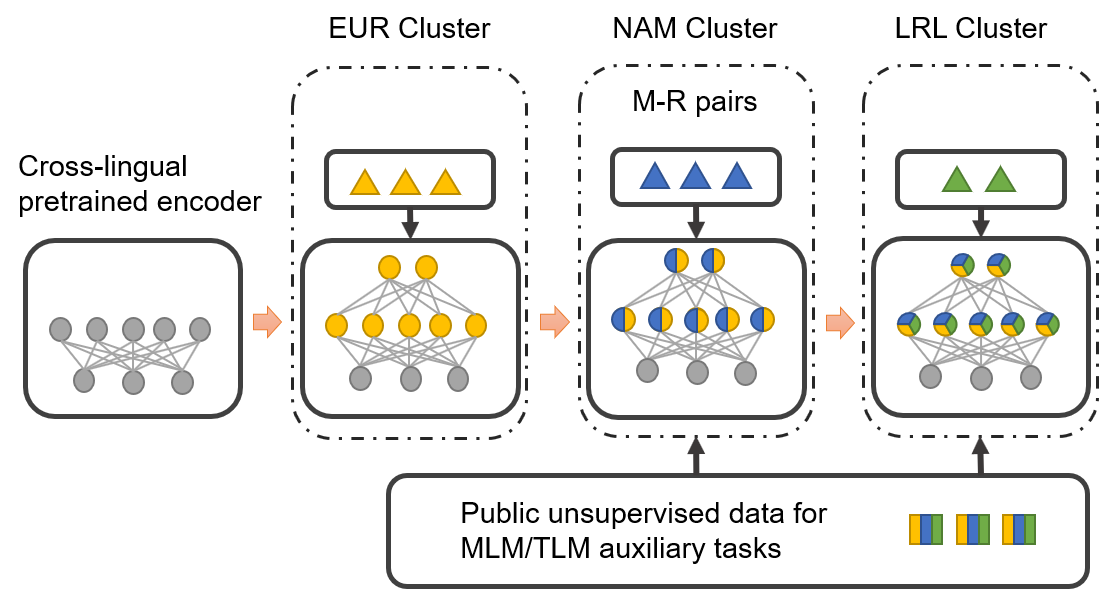}
                \label{fig:b}
                \end{minipage}
            }%
\caption{(a) Matching model architecture with symmetric loss and TLM/MLM cross-entropy loss. 
(b) Multi-task continual training loop for EUR->NAM->LRL clusters.
}
\label{fig1}
\end{figure}

IR-based model requires a fixed response set. To generate that, we collect differentially private (DP) \cite{gopi2020differentially} and anonymized replies, filtered for sensitive content from the training data which preserves user privacy while mining actual user responses. Furthermore, we use human curation to edit responses for cultural-sensitivity, gender-neutrality, etc. DP filtration requires a large amount of data due to low yields. For low-resource markets, we translate English responses with human curation for cultural adaptation to languages and locales. 

During prediction, we compute the matching score ($\cdot$) between the message and pre-computed response set vectors. Similar to \cite{HendersonASSLGK17,Deb2019DiversifyingRS}, we add a language-model (LM) penalty representing the popularity of responses to bias the predictions towards more common ones.  Translated responses inherit the penalty score from the corresponding English responses. Using this score in equation \ref{inference} we first select top $N_1$ responses, and down-select to top $N_{2}$ after deduplication using lexical clustering, before presenting to users.

% The language-model penalty score is computed with response frequency and language model score. In addition, we use lexical edit distance between responses to create response clusters to de-duplicate similar responses. 

{\small
    \begin{equation}\label{inference}
        % \begin{aligned}
        Score = \phi(m_{i}) \cdot \phi_{K}(r_{k})) + \alpha LM_{K}(r_{k})
        % \end{aligned}
    \end{equation}
}

%% file: tex/universal_model.tex
The universal SR model consists of parallel encoder architecture trained using symmetric loss function similar to the core SR model. 
%Inspired by the great improvements of transformer-based models, 
We initialize the m-r encoders with InfoXLM \cite{chi2020infoxlm}, an XLM-Roberta \cite{conneau2019unsupervised} equivalent multi-lingual model as shown in as Figure \ref{fig:a} which creates language-agnostic text representation across 100 languages. The encoder is pre-trained with both publicly available and internal proprietary corpora and has shown good cross-lingual transfer capabilities on benchmarks such as XNLI \cite{conneau2018xnli}. %,liang2020xglue}. 
%and Tatoeba sentence retrieval \cite{artetxe2019massively}.

% composed of 12 layers of Transformers and 768 hidden states, 

% We initialize the m-r encoders from the multi-lingual pre-trained model as Figure \ref{fig:b}. 
%Previous studies on task adaptation with pre-trained encoders \cite{lee2019would, peters2019tune}, show that freezing partial layers can maintain the model quality while reducing training time during fine-tuning. Following similar strategies, we freeze the embedding layer
%and $L$ bottom layers of Transformer stack during training.
% with specific tricks, which are presented in following sections.

Using a universal pre-trained model in itself enables language expansion. 
However, as we discuss next, data movement constraints made training the universal model tricky, with performance frequently worse than single mono-lingual models.
% {\color{red} --wei this claim seems very strong, can we add some citations or briefly discuss why? }
% To both overcome the data challenges as well as fully utilize the improved representation through shared learning across languages. 

\subsection{Continual Learning}

Joint training of universal encoders has led to enormous progress on standard benchmarks and industrial applications such as \citep{ranasinghe2020multilingual, gencoglu2020large}.
% We already refer to benchmarks in Section 3.

However, privacy policies restrict the data movement across geographic clusters. This prevents the joint training at a single compute cluster. As a result, we train the model sequentially in a continual learning fashion by fine-tuning the model in one region, and then continue training in another. 

The actual sequence of how this is conducted is important. We observed that keeping English at the last stage provides the best performance. This is likely because English data (which frequently contains bilingual data through code-switching) covers a large proportion in pre-training corpora, thus serving as an anchor in subsequent training stage to maintain the universal properties of the model.

% (moved from experiment section) For EUR region, we jointly fine-tune the pretrained model on all 3 languages: German, French and Italian. Similarly, for NAM region, we jointly fine-tune the pretrained model on all 3 languages: English, Spanish and Portuguese. In our experiments, we found the last stage of training containing English data is crucial to reducing catastrophic forgetting and maintain the universal properties of the model. This is likely because there pre-training corpus includes a lot of English monolingual and bilingual data. 

\subsection{Multi-task Learning}
Training the SR model in multiple stages can lead to catastrophic forgetting, where new knowledge easily supplants old knowledge. This problem can be alleviated to some extent by freezing layers of the pre-trained encoders but is still significant after the model is fine-tuned with large corpora.

% where new knowledge easily supplants old knowledge
% , while data in EUR serve as inferior anchor compared with English.
% which is plastic but unstable
Several papers have leveraged self-supervised pre-training tasks based on bi-lingual parallel corpora to create or enhance cross-lingual representations  \cite{devlin2018bert,conneau2019unsupervised,chi2020infoxlm}. Following such approaches, we experiment with Translation Language Model (TLM) \cite{lample2019cross} in continual learning to preserve the universal properties of the model. A total of $79M$ translation pairs from WikiMatrix \cite {schwenk2019wikimatrix} and MultiParaCrawl \cite{aulamo2020opustools} data including the languages considered in production are extracted as training data. In addition, we conduct an ablation study on auxiliary task selection by comparing with Masked Language Model (MLM) \cite{devlin2018bert} trained on $370M$ samples from Wikipedia.
%two auxiliary tasks: Masked Language Model (MLM) \cite{devlin2018bert} and Translation Language Model (TLM) \cite{lample2019cross}. These tasks, commonly used in self-supervised pre-training of encoders, proved equally effective at preserving the universal properties of the model during the continual learning process.

%For the MLM task, we gather $300M$ samples from Wikipedia that covers the supported languages discussed in this paper. For the TLM task, we employ a total of $79M$ translation pairs from WikiMatrix \cite {schwenk2019wikimatrix} and MultiParaCrawl \cite{aulamo2020opustools} data, including English and the supported languages.
% Wiki-MLM:369M rows
% Wiki-TLM: 15808330
% MultiParaCrawl: 63104673

The multi-task training alternates between SR and auxiliary tasks according to a set proportion of  mini-batches in an epoch. The proportion controls the trade-offs between the tasks, to achieve the desired levels of performance in the system.

% Running all the batches approximate the joint objectives of all tasks, and we can easily pick pretext tasks with native data in our target languages to enhance the multi-lingual transferability of the encoder.

% (moved from experiment) To create a single universal model, we adopted the approach of continue training the EUR model in NAM. Without any auxiliary tasks, we can see that model suffers from catastrophic forgetting resulting in inferior metrics on EUR languages. Auxiliary tasks MLM and TLM alleviate forgetting and enable continue training in a way which preserves the universal properties of the model.

%Further, we notice that TLM outperforms MLM and our hypothesis is that is because TLM uses parallel corpora and helps align semantically similar text from different languages to similar representations. 

\subsection{Data Augmentation}
Native supervised data (m-r pairs) is currently not available for low-resource languages. In such cases, English data is leveraged to generate pseudo m-r pairs using machine-translation (MT). We utilize MT data in continual learning process with auxiliary tasks, or with adapters \cite{adapters_houlsby19a} by introducing additional parameters in the transformer layers. When training with adapters, we freeze all parameters except the adapters.

% {\color{blue}
% --Qianlan: @Payal, For section 4.1-4.3: just mention the techniques applied, and for the specific usage in training loop 'like Stage 3', 'rm tlm', put it into 'Universal Model Training Loop';
% The results can be added into Table 2.
% }

% {\color{green}
% Did you implement it or use Hugging Face's or some other code? If so please clarify. 
% }

% \textit{train-translate-all} where the training data is translated to the target language and used to fine-tune the model. We use our internal machine translation model service to translate all the message and reply pairs from English to target languages to supplement the data in low-resource markets. 

% To maximize the usage of the translated data, we translate from English data stored in a cluster with longer retention permission separated from previous compute clusters. As we apply this technique in continual learning framework, we observe similar catastrophic forgetting issues as natural training data. Thus, during the loop for training the zero-resource languages, we apply multi-task learning in addition to data augmentation for fine-tuning.

\subsection{Universal Model Training Loop} 

% {\color{red} --Wei: this sections seems too specific, probably we move it under the experiments section}

% {\color{blue}
% --Qianlan: @Budha, how about moving it as first subsection in experiment?
% }

% {\color{green}
% Yes, we can do that. Just that the loop is basically the most important technical part of the paper. 
% }

The production system targets 5 high-resource languages (HRL): Spanish (ES), Portuguese (PT), French (FR), German (DE), Italian (IT) with rich native data, and 5 low-resource languages (LRL): Chinese (ZH), Japanese (JA), Dutch (NL), Czech (CS) and Hungarian (HU)  without any supervised data. English (EN) serves as pivot language in our experiments. As shown in Table \ref{tab:data}, the data is distributed across Europe (EUR), North America (NAM) and a dedicated cluster storing MT data for LRL. Data movement across these regions is not allowed. Public task-agnostic data for auxiliary tasks in 8 languages is accessible in all regions.

\begin{table}[htbp]
  \centering
  \scalebox{0.75}{
    \begin{tabular}{lll}
    \toprule
    \textbf{Region} & \textbf{Languages} & \textbf{Category} \\
    \midrule
    EUR & DE, IT, FR & High-resource \\
    \midrule
    NAM & ES, PT, EN & High-resource \\
    \midrule
    LRL & ZH, JA, NL, CS, HU & Low-resource* \\
    \bottomrule
    \end{tabular}}%
  \caption{Regional distribution of training data for different languages. *: data translated from EN. }
  \label{tab:data}%
\end{table}%
% Regional distribution of training data for different languages. *: Only MT data translated from EN.

%The first 5 High-Resource Language (HRL) data is stored scattering across Europe cluster (EUR: FR, DE, IT) and North America cluster (NAM: EN, ES, PT) following the regional policies. For low-resource languages without any native data, a separate cluster named as "Low-Resource Language (LRL) cluster" is dedicated to store MT data (ZH and JA) as well as the pivot language data (EN). Public task-agnostic data for auxiliary tasks in 8 languages is stored in all of the regions.

% - \textbf{EUR}: French, German and Italian data.

% - \textbf{NAM}: English, Spanish and Portuguese data.

% - \textbf{LRL}: machine translated Chinese, machine translated Japanese, and English data.

We train the model sequentially in 3 stages as shown in Figure \ref{fig:b}. First, we jointly train the model in EUR for FR, DE, and IT. Next, we move the model to NAM and continue train with EN, ES, and PT along with auxiliary task. Finally, in LRL, we train the model on machine translated m-r pairs along with original EN data in 2 different ways: (1) jointly train with auxiliary task, or (2) infuse the model with low-resource language adapters. In all stages, we freeze the embedding layer of the encoder during fine-tuning. According to previous studies \cite{lee2019would, peters2019tune}, freezing partial layers can maintain the model quality while reducing training time during fine-tuning. We observed that freezing embedding layer provides a good balance between micro-batch size per GPU (low if no layers are frozen) and learning capacity of the model (low if many layers are frozen).

%At each stage we alternate between the SR and TLM tasks to maintain the universality of the model and reduce over-fitting.

\subsection{Universal Model Graph for Serving}
For deployment, we create a composite graph with pre-computed response vectors of all languages embedded into the main model. A separate language identifier switches the prediction vectors to the predicted language of the input at run-time. Besides, several auxiliary models are added in online system to decide whether to trigger the universal model according to the characteristics of input message such as length and detected language. 

%% file: tex/experiments.tex
%For the pre-trained text encoder, a XLM-R like cross-lingual language model with 12 layers and 768 hidden states is applied in SR model training.

%\textbf{This section can be moved inside the universal model section without say the experimental setup of batch sizes etc.}

% We train the model in 3 stages in sequential order as below: first we train the model in \textit{Region 1} where we fine-tune the pre-trained model using \textit{SymmLoss} with French, German and Italian data. Second, we continue to train the model with data in \textit{Region 2} English, Spanish and Portuguese data with \textit{SymmLoss} and the auxiliary pre-training task with WikiMatrix\cite {schwenk2019wikimatrix} and MultiParaCrawl \cite{aulamo2020opustools} data. We experiment with both MLM and TLM tasks in the continual training. Finally, in the third stage, we move the model to \textit{Region 3} train on machine-translated Chinese and Japanese data along with the original English data. Similarly, an auxiliary task with MLM or TLM objective is considered to alleviate the forgetting issue in continual learning.

The training data is collected and processed without any eyes access from commercial users in Outlook email system. To be more specific, we filter 50M m-r pairs from one-to-one conversations for each high-resource language, and translate 20M m-r pairs for each low-resource language. Considering the m-r length distribution, we truncate m-r pairs to (96, 64) tokens as training data, and filter out messages longer than 96 tokens during inference, so that the model is more focused on providing quick responses to short messages. The response set size for each language is 20K, filtered or trans-created from English native data. 

In all three stages of training, we use an effective batch size of 16K. We utilize the Adam optimizer \cite{kingma2014adam} with weight decay and set peak learning rates as [5e-4, 3e-4, 1e-4] for three stages respectively.
We train up to 30 epochs from which the best model is selected based on validation set loss over all languages.

For MLM/TLM objectives, we use single-token masking, the task proportion is set as $0.5$. The final loss of the model is sum of symmetric loss and auxiliary task loss.
For adapters, we use the hidden dimension of 256 in the bottleneck architecture and initialize these parameters with a normal distribution of mean 0 and standard deviation 0.01. According to our observation, high standard deviation for initialization can cause divergence. All experiments are conducted with $16$ Nvidia V100-32GB GPU cards. 

During prediction, we pick top $N_1=30$ responses according to equation \ref{inference}, and then cluster the ranked results and down-select $N_2=3$ responses as final prediction.
% We creates an inference stack in run-time, in which the model receives the signal of detected language and region, and predicts with the pre-computed response encodings from specified language.  

\subsection{Offline Evaluation Metrics and Sets}
We compute evaluation metrics based on two kinds of evaluation sets. The first test set samples m-r pairs, where reply is contained in the response set (GoldenMR) and is used for computing the ranking metric, Mean Reciprocal Rank: $MRR = \frac{1}{N}\sum_{i=1}^{N}\frac{1}{Rank_i}$, for the top 15 predictions.
%, 2) Macro MRR: $MMRR = \frac{1}{R}\sum_{rsp=1}^{R}{MRR_{rsp}}$. 

The second set consists of general m-r pairs (GenMR) where the reply is not restricted to the response set. \textit{weighted}-ROUGE metrics is computed on final 3 responses with the reference response over uni/bi/tri-grams ($W\_ROUGE = \sum_{i=1}^{3}\frac{1}{w_{i}}ROUGE_{i}(Ref, Rep_{k})$), with weights of $1:2:3$ proportions.

We use $\sim$50K GoldenMR and 500K GenMR dataset for each language. For languages without native data, an evaluation proxy with MT data is used for model selection before online deployment. We give a higher preference to ROUGE as it showed higher correlation to our online metrics.

% {\small
% 	\begin{gather}
%     	MRR = \frac{1}{N}\sum_{i=1}^{N}\frac{1}{Rank_i}\\
%         MacroMRR = \frac{1}{R}\sum_{rsp=1}^{R}{MRR_{rsp}}\\
%     	W\_ROUGE = \sum_{i=1}^{3}\frac{1}{w_{i}}ROUGE_{i}(Ref, Rep_{k})
%     \end{gather}
% }
%For qualitative evaluation, we use human judgements to evaluate defect rate, duplicate rate and SBS analysis compared with individual monolingual models for each language.
%\textbf{@Qianlan: delete SBS analysis since the metrics are incomplete (only PT, ES and IT)}

\subsection{Online Evaluation Metrics}
For the deployed models in production, we measure the following online metrics on real user traffic. 

\textbf{Click-through rate (CTR)}: the ratio of the count of replied emails with SR clicks over all emails that the feature is rendered.

\textbf{Usage}: the ratio of count of replied emails with SR clicks to all replied emails. This captures the contribution of SR to all Email replies.

\textbf{Char-saved}: the average number of characters-saved by clicking the selected reply.

\subsection{Results}
The model is evaluated on the international markets we are expanding to. English is excluded as EN model is well established. 
Results on baseline (existing per-language production models) and universal models for high-resource markets are reported in Table \ref{tab:overall-native}. Results targeting new markets without any native data are reported in Table \ref{tab:overall-all}.
Entries in the tables are defined as follows:
%{\color{red} --wei:  table 1 and table 2 seem having a lot of overlap, probably we can merge these two tables}{\color{blue} --Qianlan: removed BiLSTM in table 2. @Payal, how about adding results of Adapter on first 5 or mention that they == UniPLM-HRL?; For ZH, there seems an aether pipeline issue on MacroMRR?}

\textbf{BiLSTM}: Per-language (mono-lingual) production models for non-EN markets as the baseline and also the control setting of online A/B tests. Here the encoders have shared embedding size of 320 and 2 BiLSTM layers with hidden size of 300.

%\textbf{PLM}: Per-language model with MT data (for ZH and JA) and pre-trained multi-lingal encoder. It serves as the per-language baseline model for low-resource languages in production system.

\textbf{UniPLM-[\textit{NAM/EUR}]}: Universal model created by fine-tuning pre-trained multi-lingual encoders for EUR and NAM regions respectively.

\textbf{UniPLM-HRL}: The model across the first 2 stages with the universal training loop in Figure \ref{fig:b}. In the second stage, the model is fine-tuned along with TLM auxiliary task with multi-lingual unsupervised data. This is the first universal model candidate that breaks down the data boundary across High-Resource Languages (HRL).

\begin{table}[htbp]
  \centering
   \scalebox{0.71}{
    \begin{tabular}{rrlcc}
    \toprule
    \multicolumn{1}{l}{\textbf{Reg}} & \multicolumn{1}{l}{\textbf{Lang}} & \textbf{Model} & \textbf{MRR}  & \textbf{W\_ROUGE} \\
    \midrule
    \multicolumn{1}{l}{EUR} & \multicolumn{1}{l}{DE} & BiLSTM-de & 0.3263 & 0.0685 \\
       &    & UniPLM-EUR & \textbf{0.4185} & \textbf{0.0698} \\
       %&    & UniPLM-NAM & 0.1934  & 0.0544 \\
       &    & UniPLM-HRL & 0.3323  & 0.0663 \\
\cmidrule{2-5}       & \multicolumn{1}{l}{FR} & BiLSTM-fr & 0.4569  & 0.0642 \\
       &    & UniPLM-EUR & \textbf{0.4721}  & \textbf{0.0647} \\
      % &    & UniPLM-NAM & 0.2698  & 0.0572 \\
       &    & UniPLM-HRL & 0.4135  & 0.0624 \\
\cmidrule{2-5}       & \multicolumn{1}{l}{IT} & BiLSTM-it & 0.3300  & 0.0330 \\
       &    & UniPLM-EUR & \textbf{0.4819}  & \textbf{0.0385} \\
       %&    & UniPLM-NAM & 0.2758  & 0.0314 \\
       &    & UniPLM-HRL & 0.4186  & 0.0360 \\
    \midrule
    \multicolumn{1}{l}{NAM} & \multicolumn{1}{l}{ES} & BiLSTM-es & 0.3248  & 0.0511 \\
      % &    & UniPLM-EUR & 0.1688  & 0.0397 \\
       &    & UniPLM-NAM & 0.3186  & \textbf{0.0565} \\
       &    & UniPLM-HRL & \textbf{0.3319} & 0.0552 \\
\cmidrule{2-5}       & \multicolumn{1}{l}{PT} & BiLSTM-pt & \textbf{0.4383}  & 0.0552 \\
      % &    & UniPLM-EUR & 0.0647  & 0.0385 \\
       &    & UniPLM-NAM & 0.4216  & \textbf{0.0577} \\
       &    & UniPLM-HRL & 0.4154  & 0.0563 \\
    \bottomrule
    \end{tabular}}%
  \caption{Evaluation on HRL (EUR and NAM) with UniPLM-HRL via continual multi-task learning and production baselines. The best results are in bold.}
  \label{tab:overall-native}%
\end{table}%

For new languages without native data, we continue to train the base universal model (UniPLM-HRL) with MT data with two approaches.

\textbf{UniPLM-All-CL}: The UniPLM-HRL model exported to LRL region trained with MT data (and native EN data) with SR and TLM multi-task objectives.

% Similarly as stage 2, the model is fine-tuned with TLM task in multi-task learning objective. It targets at supporting all languages including the ones without any organic data.

\textbf{UniPLM-All-ADP}: The model trained with MT-adapter, with all parameters frozen except for adapters parameters.

\begin{table}[htbp]
  \centering
  \scalebox{0.71}{
    \begin{tabular}{rrlcc}
    \toprule
    \multicolumn{1}{l}{\textbf{Reg}} & \multicolumn{1}{l}{\textbf{Lang}} & \textbf{Model} & \textbf{MRR} & \textbf{W\_ROUGE} \\
    \midrule
    \multicolumn{1}{l}{EUR} & \multicolumn{1}{l}{DE} & UniPLM-HRL & \textbf{0.3323} & 0.0663 \\
       &    & UniPLM-All-CL & 0.3103 & \textbf{0.0686} \\
    \cmidrule{2-5}       & \multicolumn{1}{l}{FR} & 
    UniPLM-HRL & 0.4135  & 0.0624 \\
       &    & UniPLM-All-CL & \textbf{0.4207} & \textbf{0.0659} \\
    \cmidrule{2-5}       & \multicolumn{1}{l}{IT} & 
    UniPLM-HRL & 0.4186 & 0.0360 \\
       &    & UniPLM-All-CL & \textbf{0.4274} & \textbf{0.0374} \\
    \midrule
    \multicolumn{1}{l}{NAM} & \multicolumn{1}{l}{ES} & UniPLM-HRL & \textbf{0.3319} & \textbf{0.0552} \\
       &    & UniPLM-All-CL & 0.3160  & 0.0551 \\
    \cmidrule{2-5}       & \multicolumn{1}{l}{PT} & 
    UniPLM-HRL & \textbf{0.4154}  & \textbf{0.0563} \\
       &    & UniPLM-All-CL & 0.3783  & 0.0561 \\
    \midrule
    \multicolumn{1}{l}{LRL} & \multicolumn{1}{l}{ZH} 
    %& PLM-zh & 0.2787  & 0.0891 \\
     %  &    
       & UniPLM-HRL & 0.1365  & 0.0740 \\
       &    & UniPLM-All-CL & 0.2638  & 0.0869 \\
       &    & UniPLM-All-ADP & \textbf{0.3024}  & \textbf{0.0901} \\
\cmidrule{2-5}       & \multicolumn{1}{l}{JA} 
    %& PLM-ja & 0.3395 & 0.1144 \\
       %&    
       & UniPLM-HRL & 0.1475 & 0.1010 \\
       &    & UniPLM-All-CL & 0.3281 & 0.1106 \\
       &    & UniPLM-All-ADP & \textbf{0.3719} & \textbf{0.1180} \\
\cmidrule{2-5}       & \multicolumn{1}{l}{NL} 
    %& PLM-ja & 0.3395 & 0.1144 \\
       %&    
       & UniPLM-HRL & 0.0638 & 0.0371 \\
       &    & UniPLM-All-CL & 0.1822 & 0.0436 \\
       &    & UniPLM-All-ADP & \textbf{0.2490} & \textbf{0.0480} \\
\cmidrule{2-5}       & \multicolumn{1}{l}{CS} 
    %& PLM-ja & 0.3395 & 0.1144 \\
       %&    
       & UniPLM-HRL & 0.0366 & 0.0386 \\
       &    & UniPLM-All-CL & 0.1312 & 0.0441 \\
       &    & UniPLM-All-ADP & \textbf{0.2612} & \textbf{0.0526} \\
\cmidrule{2-5}       & \multicolumn{1}{l}{HU} 
    %& PLM-ja & 0.3395 & 0.1144 \\
       %&    
       & UniPLM-HRL & 0.0420 & 0.0356 \\
       &    & UniPLM-All-CL & 0.0779 & 0.0776 \\
       &    & UniPLM-All-ADP & \textbf{0.2615} & \textbf{0.0907} \\
    \bottomrule
    \end{tabular}}%
   \caption{Results with UniPLM-All-CL and UniPLM-All-ADP continually augmented with MT data.}
  \label{tab:overall-all}%
\end{table}%

\subsection{Model Quality Analysis} 
% The results of our proposed universal models are listed as UniPLM-HRL and UniPLM-All-CL in the tables.
% Add aggregate metrics
Table \ref{tab:overall-native} compares the universal model UniPLM-HRL with both per-language baselines and per-region models. Table \ref{tab:overall-all} shows the results with the low-resource languages, which are trained with data augmentation approach involving MT data, with multi-task learning or adapters.

\textbf{Per-language vs. Universal Model}: The BiLSTM production models serve as strong baselines and have comparable MRR for UniPLM-NAM in ES and PT (Table \ref{tab:overall-native}). UniPLM-EUR has better performance than the BiLSTM production models. Overall, the Uni-PLM models have comparable or better performance than the monolingual baselines. 

% jointly trained high resource languages has better performance than the BiLSTM baselines.

% By comparing UniPLM-EUR with each BiLSTM per-language model in EUR, and similarly for UniPLM-NAM versus BiLSTM model in NAM, universal model jointly trained on languages with enough data shows superior performance and serves as upper bound in current model building paradigm.

\textbf{UniPLM-NAM/EUR vs. UniPLM-HRL}: 
Table \ref{tab:overall-native} also shows no appreciable difference in ROUGE metrics when training the model in 2 stages. In addition, the model outperforms BiLSTM per-language models on MRR on ES, DE, FR, and IT. 

The above two comparisons show that for high-resource languages, we do not suffer significant degradation in quality with single stage and two-stage universal models. 

% presents a close gap of $\sim$0.002 point between UniPLM-HRL and UniPLM-\textit{Region} models in terms of \textit{ROUGE} in EUR and NAM, respectively. No appreciable difference is observed on all metrics compared with the per-region trained universal models. 

% This confirms the effectiveness of our modeling approach to maintain universal properties. 

% From another perspective by comparing UniPLM-NAM and UniPLM-HRL in EUR, both models are fine-tuned with same corpora in NAM at the last stage. However, the universal model successfully surmounts the obstacle of data boundary to retain nearly all of the improvement learning from cross-region data at preceding stage, and achieves significant gain over UniPLM-NAM. 

\textbf{Performance on LRL}:
Table \ref{tab:overall-all} compares the UniPLM-All-CL and UniPLM-All-ADP with UniPLM-HRL model on low-resource languages. While UniPLM-HRL shows poor ranking performance, UniPLM-All-CL significantly improves on all metrics for LRL, while preserving the ROUGE performance on the other 5 languages. With adapters, UniPLM-All-ADP outperforms other models on all metrics in low-resource languages while keeping the performance unchanged (as a result of freezing the UniPLM-HRL model) in both EUR and NAM.

% We observe that continual training on MT data slightly degrades performance on EUR and NAM languages.
Overall, the results demonstrate the effectiveness of MT data augmentation in low-resource languages. We observe slight performance degradation on EUR and NAM languages caused by continual training on MT data. This may be due to imperfect translation. However we can mitigate these losses with MT-adapters which are quite promising as they increase the parameters by just $4.3\%$ and even improves training efficiency as we can freeze all other parameters during fine tuning.
% Another advantage of adapter-based model is that as we are fine-tuning only the adapter layers with a significant improvement on training efficiency.

% Table generated by Excel2LaTeX from sheet 'Overall Results'
\begin{table}[htbp]
  \centering
    \scalebox{0.71}{
    \begin{tabular}{rrlcc}
    \toprule
    \multicolumn{1}{l}{\textbf{Reg}} & \multicolumn{1}{l}{\textbf{Lang}} & \textbf{Model} & \textbf{MRR} & \textbf{W\_ROUGE} \\
    \midrule
    \multicolumn{1}{l}{EUR} & \multicolumn{1}{l}{DE} &  UniPLM-HRL & 0.3323  & 0.0663  \\
       &    & \multicolumn{1}{r}{-TLM}& \textbf{0.3643}  & \textbf{0.0701} \\
       &    & \multicolumn{1}{r}{-TLM+MLM} & 0.3070  & 0.0596 \\
\cmidrule{2-5}       & \multicolumn{1}{l}{FR} &  UniPLM-HRL & \textbf{0.4135}  & \textbf{0.0624}  \\
       &    & \multicolumn{1}{r}{-TLM}& 0.3772  & 0.0583 \\
       &    & \multicolumn{1}{r}{-TLM+MLM} & 0.4126  & 0.0606 \\
\cmidrule{2-5}       & \multicolumn{1}{l}{IT} &  UniPLM-HRL & 0.4186  & \textbf{0.0360}  \\
       &    & \multicolumn{1}{r}{-TLM}& \textbf{0.4284} & 0.0359 \\
       &    & \multicolumn{1}{r}{-TLM+MLM} & 0.4035 & 0.0343 \\
    \midrule
    \multicolumn{1}{l}{NAM} & \multicolumn{1}{l}{ES} & UniPLM-HRL & \textbf{0.3319} & \textbf{0.0552}   \\
       &    & \multicolumn{1}{r}{-TLM} & 0.2958  & 0.0543 \\
       &    & \multicolumn{1}{r}{-TLM+MLM} & 0.3023  & 0.0537 \\
\cmidrule{2-5}       & \multicolumn{1}{l}{PT} & UniPLM-HRL & 0.4154  & \textbf{0.0563}   \\
       &    & \multicolumn{1}{r}{-TLM}& 0.4176  & 0.0561 \\
       &    & \multicolumn{1}{r}{-TLM+MLM} & \textbf{0.4234}  & 0.0559 \\
    \bottomrule
    \end{tabular}%
    }
    \caption{Results with variations on UniPLM-HRL. -TLM denotes removing TLM and -TLM+MLM denotes replacing with MLM in continual learning. }
  \label{tab:auxiliary_task}%
\end{table}%

\begin{table}[htbp]
  \centering
  \scalebox{0.71}{
    \begin{tabular}{rrlcc}
    \toprule
    \multicolumn{1}{l}{\textbf{Reg}} & \multicolumn{1}{l}{\textbf{Lang}} & \textbf{Model} & \textbf{MRR} & \textbf{W\_ROUGE} \\
    \midrule
    \multicolumn{1}{l}{EUR} & \multicolumn{1}{l}{DE} & UniPLM-HRL& 0.3323 & 0.0663 \\
       &    &  \multicolumn{1}{r}{+EUR} & 0.4272 & 0.0708 \\
\cmidrule{2-5}       & \multicolumn{1}{l}{FR} & UniPLM-HRL& 0.4135  & 0.0624 \\
       &    & \multicolumn{1}{r}{+EUR} & 0.4818 & 0.0660 \\
\cmidrule{2-5}       & \multicolumn{1}{l}{IT} & UniPLM-HRL& 0.4186  & 0.0360 \\
       &    & \multicolumn{1}{r}{+EUR} & 0.4851 & 0.0388 \\
    \midrule
    \multicolumn{1}{l}{NAM} & \multicolumn{1}{l}{ES} & UniPLM-HRL& 0.3319  & 0.0552 \\
       &    & \multicolumn{1}{r}{+EUR} & 0.2125  & 0.0456 \\
\cmidrule{2-5}       & \multicolumn{1}{l}{PT} & UniPLM-HRL& 0.4154  & 0.0563 \\
       &    & \multicolumn{1}{r}{+EUR} & 0.3298  & 0.0505 \\
    \bottomrule
    \end{tabular}%
    }
  \caption{Results with 2-stage and replay-based continual learning. +EUR denotes replaying UniPLM-HRL with EUR m-r pairs.}
  \label{tab:continual}%
\end{table}%

% Based on UniPLM-HRL, we conduct the study to investigate on the contributions of auxiliary tasks and continual learning sequence.
% , and we also want to understand (3) whether multi-lingual pre-trained model contributes to universal modeling with enough training data.
\subsection{Ablation Studies}
\textbf{MLM and TLM auxiliary tasks}:
Table \ref{tab:auxiliary_task} investigates contributions of auxiliary tasks in UniPLM-HRL model. We remove TLM objective as \textit{-TLM} which represents continue training only on SR task, and replace TLM with MLM objective as \textit{-TLM+MLM} which represents joint training with SR and MLM tasks. UniPLM-HRL with TLM task shows improvements over MLM task and also outperforms single SR task for W\_ROUGE for all languages except DE. We hypothesize that TLM uses bi-lingual corpora which helps align representations for semantically similar text from different languages in task-specific fine-tuning. Furthermore, TLM objective can be interpreted as maximizing mutual information between cross-lingual contexts implicitly \cite{chi2020infoxlm}. It demonstrates that such inductive biases in auxiliary tasks are important for cross-lingual transfer in universal models.

\textbf{Replay in continual learning}: We continue to train the UniPLM-HRL model by rehearsing the old data in EUR as \textit{+EUR}. In Table \ref{tab:continual}, \textit{+EUR} we see severe regression on NAM languages, despite the improvement on EUR languages. The \textit{replay} concept in continual learning \cite{mcclelland1998complementary} fails here due to the two reasons. First, forgetting is the quintessential mode of continual learning. Second, EUR iteration doesn't contain the pivot language English training data. Continual learning requires delicately maintaining the universal properties through knowledge anchors which is difficult to achieve in practice.
%I think we should mention that EUR iteration doesn't contain the pivot/anchor language EN which can be one of the reasons

\subsection{Online Results}
Based on the offline metrics, we selected UniPLM-HRL as the first candidate for online tests in our production system. 
Using BiLSTM per-language model as the control, we conducted a 2-week A/B test with 5\% user traffic for each model per language/region. Table \ref{tab:online} presents the results for different languages. We observe statistically significant gain in ES (CTR) and FR (Char-saved). While there are regressions in other languages, they are not statistically significant ($p>0.5$)

% To be noted, nearly 30\% of overall Char-saved in 5 languages and 7 regions is reduced. No significant regression (>0.05 p-val) is observed.

% Table generated by Excel2LaTeX from sheet 'Overall Results'
\begin{table}[htbp]
  \centering
   \scalebox{0.73}{
    \begin{tabular}{lccc}
    \toprule
    \textbf{Lang} & \textbf{CTR} (p-val) & \textbf{Usage} (p-val) & \textbf{Char-saved} (p-val) \\
    \midrule
    ES & \textbf{4.20\%} (0.0163) & \textbf{7.71\%} (0.0001) & -0.91\% (0.6592) \\
    PT & 0.00\% (0.3690) & 0.00\% (0.3264) & 3.32\% (0.2737) \\
    FR & -3.51\% (0.1636) & -3.14\% (0.2773) & \textbf{5.08\%} (0.0495) \\
    IT & -3.62\% (0.4506) & -7.59\% (0.1515) & 7.03\% (0.0602) \\
    DE & 2.80\% (0.5147) & -1.07\% (0.8233) & 5.39\% (0.1193) \\
    \bottomrule
    \end{tabular}}%
    \caption{Online metrics for UniPLM-HRL model. The control model is BiLSTM in each language. The numbers with p-val < 0.05 are in bold.}
  \label{tab:online}%
\end{table}%

Overall, the universal model is generally better or at par compared to their mono-lingual baselines. This has allowed us to deploy the universal model to 100\% of users in the 5 languages. An extended universal model supporting low-resource languages is getting deployed during the writing of this paper.

Compared with per-language separate model building, the effort of model training, inference stack and deployment can be substantially reduced, though the process of training data and response collection, and human evaluation for all our targeted languages are still required. Overall, around 65\% training and performance improvement time cost can be saved with one single universal model target at 5 languages. We expect even higher amortized serving costs reductions as the approach is scaled to more languages.

%We are also actively working on improving the metrics of UniPLM-All-CL model for further deployment.

% \begin{table*}[!ht]
%     \centering
%         {\small
%             \begin{tabular}{ |c|c|c|c|c|c| } 
%                  \hline
%                  Language & Model & MRR & MacroMRR & P@3 & ROUGE \\ 
%                  \hline
%                  \multirow{1}{4em}{German} & EUR + NAM w/ TLM + MT w/ TLM & 0.3103 & 0.1886 & 0.3648 &  0.0686 \\ 
%                  \hline
%                  \multirow{1}{4em}{French} & EUR + NAM w/ TLM + MT w/ TLM & 0.4207 & 0.0940 & 0.5146 & 0.0659 \\ 
%                  \hline
%                  \multirow{1}{4em}{Italian} & EUR + NAM w/ TLM + MT w/ TLM & 0.4274 & 0.1971 & 0.5465 & 0.0374 \\ 
%                  \hline
%                  \multirow{1}{4em}{Spanish} & EUR + NAM w/ TLM + MT w/ TLM & 0.3160 & 0.0991 & 0.3698 &  0.0551 \\ 
%                  \hline
%                  \multirow{1}{4em}{Portuguese} & EUR + NAM w/ TLM + MT w/ TLM & 0.3783 & 0.1392 & 0.4522 & 0.0561 \\ 
%                  \hline
%                  \multirow{2}{4em}{Chinese} & CH MT & 0.3812 & 0.0992 & 0.4777 & 0.0553\\
%                  & EUR + NAM w/ TLM + MT w/ TLM & 0.3346 & 0.0559 & 0.4128 & 0.0539 \\ 
%                  \hline
%                  \multirow{2}{4em}{Japanese} & JA MT & 0.4188 & 0.0948 & 0.4872 & 0.0578 \\
%                  & EUR + NAM w/ TLM + MT w/ TLM & 0.3612 & 0.0522 & 0.4192 & 0.0585 \\ 
%                  \hline
%             \end{tabular}
%         }
% \end{table*}